\def\year{2020}\relax
\documentclass[letterpaper]{article} 
\usepackage{aaai20}  
\usepackage{times}  
\usepackage{helvet} 
\usepackage{courier}  
\usepackage[hyphens]{url}  
\usepackage{graphicx} 
\usepackage{graphicx}  
\usepackage{color}

\usepackage{caption}
\usepackage[ruled]{algorithm2e}
\usepackage{booktabs}       
\usepackage{graphicx}
\usepackage{placeins}

\title{Disentangling GAN with\\One-Hot Sampling and Orthogonal Regularization}

\author{
  Bingchen Liu, Yizhe Zhu, Zuohui Fu, Gerard de Melo, Ahmed Elgammal \\
  Department of Computer Science\\
  Rutgers University\\
  \texttt{$\{$bingchen.liu, yizhe.zhu, zuohui.fu$\}$@rutgers.edu}
}

\usepackage{amsmath,amssymb}

\DeclareMathOperator{\E}{\mathbb{E}}
\DeclareMathOperator{\x}{\mathbf{x}}
\DeclareMathOperator{\z}{\mathbf{z}}

\begin{document}

\maketitle

\begin{abstract}
 Exploring the potential of GANs for unsupervised disentanglement learning, this paper proposes a novel GAN-based disentanglement framework with One-Hot Sampling and Orthogonal Regularization (OOGAN). While previous works mostly attempt to tackle disentanglement learning through VAE and seek to implicitly minimize the Total Correlation (TC) objective with various sorts of approximation methods, we show that GANs have a natural advantage in disentangling with an alternating latent variable (noise) sampling method that is straightforward and robust. Furthermore, we provide a brand-new perspective on designing the structure of the generator and discriminator, demonstrating that a minor structural change and an orthogonal regularization on model weights entails an improved disentanglement. Instead of experimenting on simple toy datasets, we conduct experiments on higher-resolution images and show that OOGAN greatly pushes the boundary of unsupervised disentanglement.
\end{abstract}

\section{Introduction}

A disentangled representation is one that separates the underlying factors of variation such that each dimension exclusively encodes one semantic feature \cite{bengio2013representation,kim2018disentangling}. While the benefits of the learned representation for downstream tasks is questioned by \citeauthor{locatello2018challenging}~\shortcite{locatello2018challenging}, disentangling a Deep Neural Network (DNN) is still of great value in terms of human-controllable data generation, data manipulation and post-processing, and increasing the model interpretability. Moreover, disentanglement learning in an unsupervised manner can effectively highlight the biased generative factors from a given dataset, and yield appealing data-analytic properties. In this work, we focus on the unsupervised disentanglement learning using GANs \cite{goodfellow2014generative} on images, which brings substantial advancement in tasks such as semantic image understanding and generation, and potentially aids research on zero-shot learning and reinforcement learning \cite{bengio2013representation,higgins2017beta,lample2017fader,zhu2018generative,zhu2019learning,8334453,elgammal2017can,elgammal2018shape}.

The most popular methods to tackle the unsupervised disentangling problem are based on GANs \cite{goodfellow2014generative} or VAEs \cite{kingma2013auto}, and many instantiations of these \cite{saxe2018information,achille2018information,gabrie2018entropy,alemi2018an,louizos2017bayesian} draw on information-theoretical concepts \cite{shannon1948mathematical}. InfoGAN \cite{chen2016infogan} seeks to maximize a \textbf{Mutual Information} (MI) lower bound between a sampled conditional vector and the generated data, with the expectation that the generator and discriminator will disentangle the vector with respect to the true underlying factors. In contrast, VAE-based approaches \cite{NIPS2018_7527,kim2018disentangling,esmaeili2019structured} attempt to optimize a \textbf{Total Correlation} (TC) \cite{watanabe1960information} objective imposed on the inferred latent vector, which achieves disentanglement by encouraging inter-dimensional independence in the latent vector.

TC-based VAE models have proven fruitful in disentangling. However, there is usually a trade-off between the degree of achievable disentanglement and the data-generating ability of VAE \cite{kim2018disentangling}. In practice, VAE struggles significantly when trained on higher-resolution images due to its restricted generative power. Furthermore, it only approximates the TC, since both the marginal distribution of the learned latent representation and the product of its marginals are intractable in VAE, which makes the optimization process implicit and complicated. In contrast, with rapid advances in recent years \cite{zhang2018self,miyato2018spectral,karras2018style}, GANs have become more stable to train, and their generative power has become unparalleled even on high-resolution images. Nonetheless, less attention has been paid to GANs in unsupervised disentanglement learning. Accordingly, we propose OOGAN, a novel framework based on GANs that can explicitly disentangle while generating high-quality images. The framework's components can readily be adopted to other GAN models.

Unlike in VAEs, where a latent vector has to be inferred, in GANs, noise is actively sampled as the latent vector during training. We exploit this property to enable OOGAN to directly learn a disentangled latent vector, by means of one-hot vectors as latent representation to enforce exclusivity and to encourage each dimension to capture different semantic features. This is achieved without sacrificing the continuous nature of the latent space through an alternating sampling procedure. We argue that our proposed OOGAN fully highlights the structural advantage of GANs over VAEs for disentanglement learning, which, to our knowledge, has not been exploited before.

We achieve disentanglement in OOGAN through three contributions: 1) We propose an alternating \textit{one-hot sampling} procedure for GANs to encourage greater disentanglement. 2) We adopt an orthogonal regularization on the model weights to better accompany our objective. 3) We identify a weakness in InfoGAN and related models with similar structure, which we summarize as the \textit{compete and conflict issue}, and propose a model-structural change to resolve it. Moreover, we propose a compact and intuitive metric targeting the disentanglement of the generative part in the models. We present both quantitative and qualitative results along with further analysis of OOGAN, and compare its performance against VAEs and InfoGAN.

\section{Related Work}

\textbf{$\beta$-VAE-based models:} In the settings of $\beta$-VAE and its variants \cite{higgins2017beta,burgess2018understanding,NIPS2018_7527,kim2018disentangling,esmaeili2018structured}, a factorized posterior $p_\phi(\z|\x)$ is learned such that each dimension of a sampled $z_i$ is able to encode a disentangled representation of data $\x$. The fundamental objective that $\beta$-VAE tries to maximize (also known as the Evidence Lower-Bound Optimization) is:
\begin{align}\label{eq:beta-vae}
\begin{aligned}
    \mathcal{L}(\theta, \phi; \x,\z,\beta) = \E_{q_\phi(\mathbf{z}|\mathbf{x})}&[\log{p_\theta (\x|\z)}] - \\
    &\beta D_\mathrm{KL}(q_\phi(\z|\x) || p(\z)),
\end{aligned}
\end{align}
where $\beta>1$ is usually selected to place stronger emphasis on the KL term for a better disentanglement learning. \citeauthor{burgess2018understanding}~\shortcite{burgess2018understanding} motivate the effect of $\beta$ from an information-theoretical perspective, where the KL divergence term can be regarded as an upper bound that forces $q(z)$ to carry less information, thus becoming disentangled. 

Follow-up research extends the explanation by deriving a Total Correlation from the KL term in the $\beta$-VAE objective, and highlights this TC term as the key factor to learning disentangled representations. Given a multi-dimensional continuous vector $z$, the TC  quantifies the redundancy and dependency among each dimension $z_i$. It is formally defined as the KL divergence from the joint distribution $q(z_1,...,z_n)$ to the independent distribution of $q(z_1)q(z_2)...q(z_n)$:
\begin{align}\label{eq:beta-vae}
   \mathcal{L}_\mathrm{TC} = D_\mathrm{KL}(q(z) || \hat{q}(z)) ,
\end{align}
where $\hat{q}(z) = \prod_{i=1}^{n} q(z_i)$.  
However, the TC term requires the evaluation of the density $q(z) = \E_{p(n)}[q(z|n)]$, which depends on the distribution of the entire dataset and usually is intractable. For the sake of a better optimization on the TC term, \citeauthor{NIPS2017_7178}~\shortcite{NIPS2017_7178} propose TC-VAE, which uses a minibatch-weighted sampling method to approximate TC. \citeauthor{kim2018disentangling}~\shortcite{kim2018disentangling} perform the same estimation using an auxiliary discriminator network in their Factor-VAE. Furthermore, \citeauthor{esmaeili2018structured}~\shortcite{esmaeili2018structured} suggest a more generalized objective where the marginals $q(z_i)$ can be further decomposed into more TC terms, in case each $q(z_i)$ learns independent but entangled features, which leads to a hierarchically factorized VAE. \citeauthor{dupont2018learning}~\shortcite{dupont2018learning} leverage the Gumbel Max trick \cite{jang2016categorical} to enable disentangled learning of discrete features for VAE.

\noindent\textbf{GAN-based models:} InfoGAN \cite{chen2016infogan} reveals the potential of Generative Adversarial Networks \cite{goodfellow2014generative} in the field of unsupervised disentanglement learning. In a typical GAN setting, a generator $G$ and a discriminator $D$ are trained by playing an adversarial game formulated as:
\begin{align}\label{eq:GAN}
\begin{aligned}
    \min_{G}\max_{D}\,&\mathcal{L}_\text{GAN}(D,G) = \\
    &\E_{p(x)}[\log(D(x))] + \E_{p(z)}[\log(1-D(G(z))].
\end{aligned}
\end{align}
While this mini-max game guides $G$ towards generating realistic $x$ from noise $z$ drawn from the isotropic Gaussian distribution, the variation of $z$ often remains entangled. InfoGAN manages to make $G$ learn a disentangled transformation from a latent code $c$, which is concatenated to $z$ before being fed to $G$. InfoGAN achieves this by maximizing a Mutual Information (MI) lower-bound between $c$ and the generated sample $x=G(z,c)$, where the MI $I(c, G(z,c))$ can be calculated directly by matching $c$ to $\hat{c}=Q(G(z,c))$, where $Q$ is an auxiliary network that seeks to predict the sampled latent vector from $x$. In practice, $Q$ shares most weights with $D$. However, such a lower-bound constraint only ensures $c$ gains control over the generation process, but cannot guarantee any disentanglement as $c$ increases its dimensionality, because this lower-bound does not encourage any independence across each dimension of $c$. 

A more recent GAN-based disentanglement work is the Information-Bottleneck-GAN \cite{jeon2019ibgan}. However, it fails to take advantage of the GAN structure, instead trying to implicitly minimize the TC in the same way as $\beta$-VAE. The method requires an extra network that encodes noise $z$ into to a control vector $c$ and lets the original $G$ and $D$ play the decoder's role to reconstruct $z$. This severely hurts the generation quality, since $G$ starts the generation from $c$, which has a much lower dimensionality than $z$, and the increased network modules and loss objectives make the training scheme tedious and less likely to find the proper hyper-parameters that allow the model to converge.

\begin{figure}
\centering
  \includegraphics[width=1\linewidth,height=3cm]{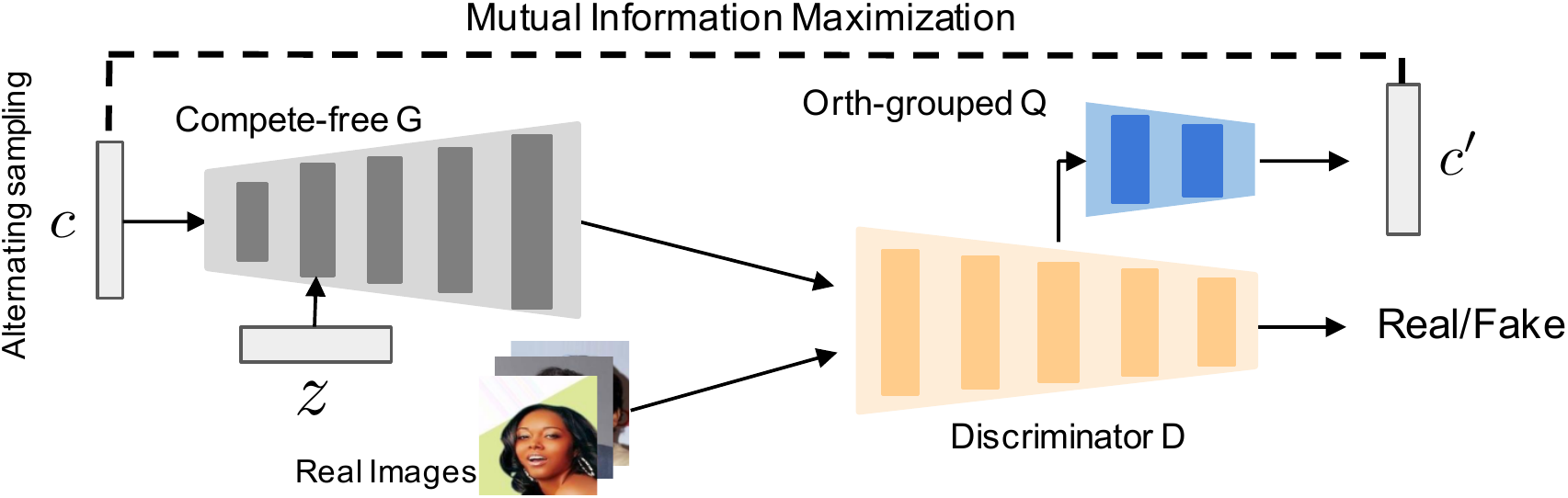}
  \caption{OOGAN makes minimal changes upon a basic GAN. $c$ denotes the continuous control vector, $z$ is the noise vector, $c'$ is the feature representation of fake images.}
  \label{fig:over_all_model}
\end{figure}

\section{Proposed Method}
Our approach accomplishes both the task of disentangled feature extraction and human-controllable data generation in an unsupervised setting within the GAN framework. We define our problem as follows: For a continuous control vector $c$ sampled from $\mathsf{uniform}(0,1)$, we wish our generator $G$ to be disentangled such that each dimension in $c$ solely controls one feature of the generated data $x=G(c,z)$ ($z$ is the noise vector), and our feature extractor $Q$ (mostly the discriminator $D$ with a few layers on top that gives vector outputs) is able to emit a feature representation $c'$, given $x$, that is disentangled in the same way as $c$. 

Our model is illustrated in Figure~\ref{fig:over_all_model}, and the complete training process is described in Algorithm~\ref{alg:1}. Similar to the design of InfoGAN, we let the feature extractor $Q$ be a sub-module that shares weights with the discriminator $D$. $Q$ takes the feature map of a generated image $G(c,z)$ as input and tries to predict the control vector $c$ used by the generator $G$. We describe the three components of our OOGAN framework in the following sub-sections.

\subsection{Alternating Continuous and One-hot Sampling}
Previous methods of minimizing TC to achieve disentanglement have two limitations. First, due to the intractability, extra network modules and objectives have to be invoked to approximate TC, which leads to undesired hyper-parameter tuning, a non-trivial training regime, and a high computational overhead. Second, to optimize the derived TC objectives in VAE-based models, the data generation quality is sacrificed \cite{kim2018disentangling,NIPS2018_7527}, and can hardly perform well on higher-resolution image data. In contrast, in the GAN setting, the latent vector is sampled instead of inferred as in VAEs. This motivates us to approach disentanglement by deliberately sampling latent vectors that possess the property of inter-dimensional independence and training the networks using these sampled vectors. 

To this end, we propose an alternating continuous-discrete sampling procedure: we alternate between sampling continuous $c$ from $\mathsf{uniform}(0,1)$ (as typically done in InfoGAN) and sampling $c$ as one-hot vectors. The one-hot vector $c$ implies that the generated image should only exhibit one feature, and, ideally,  the prediction $c'$ from $Q$ should also be a one-hot vector. On both the $G$ and $Q$ sides, any presence of other features should be penalized, while alternating with continuous uniform sampling is necessary to ensure the continuity of the representation. Interestingly, such a one-hot sampling resembles a classification task. Therefore, we can jointly train $Q$ and $G$ directly via a cross-entropy loss. In such a process, $G$ is trained to generate images that possess the specified features and avoids retaining any other features, while $Q$ is trained to summarize the highlighted feature only in one dimension and refrain from spreading the feature representation into multiple dimensions. 

Note that we treat $c$ as a continuous vector in the whole training process, and the alternating one-hot sampling can be seen as an regularizor for $G$ and $Q$. When we sample $c$ from $\mathsf{uniform}(0, 1)$ as in InfoGAN, we ensure the correlation between $c$ and $x$ remains. Furthermore, when we interleave that with one-hot samples, the process can be interpreted as getting the extremely typical samples (those samples that lie on the boundary of the uniform distribution) for the model. We argue that sampling data at the distribution boundaries makes the model pay more attention to these boundaries, yielding a clearer distribution shape highlighting the semantics of these boundary factors. These typical samples are vital for the model to learn inter-dimension exclusivity, as a one-hot $c$ regularizes $G$ to generate images with only one factor and $Q$ to only capture this one factor. 

In other words, alternating one-hot sampling and uniform sampling results in a more desirable prior distribution for disentangling GANs, which provides much more typical samples on the margin than a single uniform distribution. Such an alternating procedure, which injects the categorical sampling (i.e., the one-hot sampling) into a continuous $c$, makes it possible that $c$ gains continuous control over the generation process while simultaneously paying more attention to those typical examples, and therefore achieves better disentanglement.

Formally, our complete objective for OOGAN is:
\begin{align}\label{eq:InfoGAN}
\begin{aligned}
     \min_{G}\max_{D}\,&\mathcal{L}_\text{OOGAN}(D,G)=\mathcal{L}_\text{GAN}(D,G) + \\
    &\lambda I(c_\text{continuous}, G(c_\text{continuous}, z)) + \\
    &\gamma \mathcal{L}_\text{Cross-Entropy}(Q(G(c_\text{one-hot(d)}, z)), c_\text{one-hot(d)}).
\end{aligned}
\end{align}
Despite the lack of any TC terms in our objective, the one-hot sampling still ensures that we have a well-disentangled feature extractor $Q$ and generator $G$ that learn features with no overlap between each dimension in $c$, without any approximations and extra network modules involved.

\subsection{Compete-Free Generator}
\label{sec:g}
InfoGAN \cite{chen2016infogan} and many conditional-GAN variants leverage an auxiliary vector $c$ that is concatenated with noise $z$ before being fed into $G$, with the expectation that $c$ carries the human-controllable information. From a size perspective, the dimensionality of $z$ is usually much more significant than $c$ ($z$ typically has around a hundred dimensions, while $c$ has in the order of 10). Intuitively, $c$ will have much less impact in the generation process. With the objective of the unsupervised disentanglement learning, the large portion of influence $z$ is taking in the generation process is undesirable, which we refer to as the \textit{\textbf{competing and conflicting issue}}.

\begin{figure}
\centering
  \includegraphics[height=4.3cm, width=\linewidth]{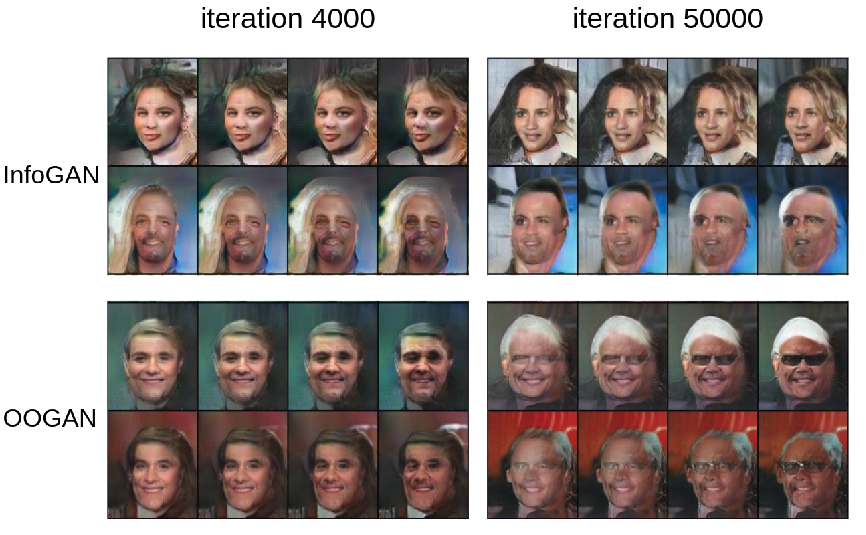}
  \caption{Latent traversals trained on CelebA to showcase the \textit{competing and conflicting issue}. The images are from the same set of ($z$, $c$) on one fixed dimension of $c$ after different training iterations. We observe that InfoGAN begins to capture what appears to be a ``wearing glasses" feature at a very early stage, but discards it during training in all dimensions of $c$. In contrast, when OOGAN begins to capture this feature, it consistently masters it in the end.}
  \label{fig:comt-free-glass}
\end{figure}

Usually, a disentangled feature learned by $c$ can also be entangled in $z$. During the training process, if $c$ with $c_i$ holding a high signal on a certain feature is paired with some $z$ with many dimensions holding the same feature with a conflict signal, this signal, entangled in $z$, will easily overpower $c$. Thus, the generated images will not present $c_i$'s signal. Such a conflict will discourage $c$ from mastering the learned feature and cause it to stray away to some easier-to-achieve but less distinct features. An example is shown in Figure~\ref{fig:comt-free-glass}. More discussion can be found in the experiments section.

To avoid the aforementioned \emph{competing and conflicting issue}, we propose a new \textbf{Compete-Free Design} of the generator's input block, which switches the role between $c$ and $z$ by letting $c$ control the fundamental content even when the dimensionality of $c$ is low, and ensures that $z$ has limited influence in the generation process. 

To start with, we project the low-dimensional control vector $c$ into a multi-channel $4\times4$ feature map by a convTransposed layer. Then, we add this feature map to a learned constant tensor with the same dimensionality. 

The weights for the constant tenor are randomly initialized before training and will be used for all generations. These weights are trained via back-propagation just like all other model weights. This learned constant can be regarded as an additive bias that is learned from the dataset, and is necessary since it is responsible for representing the features that are not captured in $c$. Ideally, when given a $c$ with all zeros as input, this constant should let the generator output the most ``neutral" $x$. In our experiments, we find such constant important for providing a stabilized learning process. It makes OOGAN faster to converge to the disentangled factors. Intuitively, one can imagine this constant as placing an anchor at the center of the target distribution, such that all latent factors can expand towards different directions. This behavior encourages the model to focus on learning the correlation between $c$ and the generated images. Without this "constant", OOGAN will still work as it is but will be slower to converge for $c$.

\begin{figure}
  \centering
  \includegraphics[width=\linewidth, height=2.6cm]{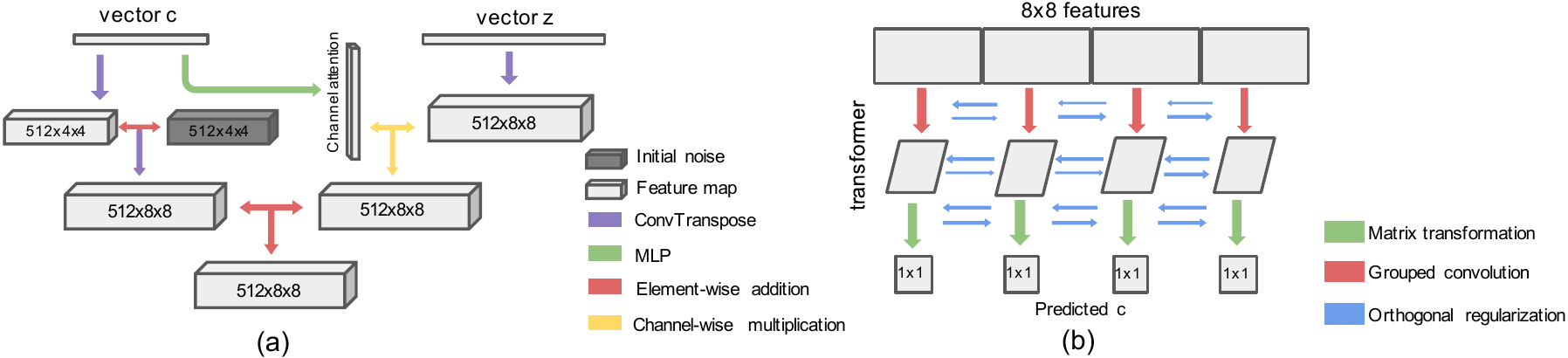}
  \caption{Model structures: (a) Input block of the compete-free $G$. (b) Orthogonal-regularized grouped $Q$ }
  \label{fig:g_and_d}
\end{figure}

To encourage the variance and complement the details for a higher-quality generation, the traditional noise $z$ is still taken into the generator, but only after the $8\times8$ feature map level. To prevent $z$ from causing the \textit{competing and conflicting issue}, we leverage an attention mask generated from $c$ on the features from $z$, which means that only the approved part of $z$ by $c$ can join in the generation process. Different layers in CNNs have been studied extensively \cite{karras2018style}, where the first few layers tend to generate fundamental compositions, and higher layers only refine the details. So our design makes $c$ more natural to control the key generative factors without the interference of $z$. The design details are illustrated in Figure~\ref{fig:g_and_d}-(a).

Our generator design resembles the one proposed for StyleGAN \cite{karras2018style}, as we both base the generation on a fixed multi-dimensional feature map instead of an input vector $z$, and take $z$ as input only in later layers. As claimed by Karras \emph{et al.}, such a design leads to a better separation in the data attributes and a more linear interpolation along latent factors. However, both the motivation and structure details are different. The \textit{disentanglement} we study here is a more strict term than what Karras \emph{et al.}\ used. The fundamental difference is that the fixed weights in our proposed model only serve as a supportive bias, and will be directly changed by $c$, while in StyleGAN the fixed weights are solely used to start the image generation process.

\subsection{Orthogonal Regularized \& Grouped Feature Extractor} 
\label{sec:d}
To learn a disentangled representation, we propose a new structure of $Q$ that uses grouped convolutions \cite{krizhevsky2012imagenet,zhang2018shufflenet} instead of traditional fully connected ones, with an orthogonal regularization on the weights among every convolution kernel. The intuition is, since we hope that $Q$ will be a highly disentangled feature extractor, a fully connected (FC) design is not favorable, since, in a FC convolution, each feature prediction has to take into consideration all the feature maps from the previous layer. A grouped convolution, on the other hand, can focus its decision making on a much smaller group of previous features, and may thus be less distracted by potentially irrelevant features. 

To make sure that each group is indeed attending to different features, we impose an additional loss function on the weights of the convolutional layers to enforce the orthogonality between different kernels. Weight orthogonality in DNNs has been studied \cite{brock2016neural,huang2018orthogonal,bansal2018can}. However, these studies each focused on different tasks, and none of them revealed the potential for disentanglement learning. 

The orthogonal regularization we use is straightforward: during each forward pass of the OOGAN, compute and minimize the cosine similarity between every convolutional kernel. With grouped feature extraction and orthogonal regularization, $Q$ structurally more easily captures diversified features in each dimension. Note that the group design is not only applicable to convolutional layers but also to grouped linear layers or other weights indicated as ``transformer" in Figure~\ref{fig:g_and_d}-(b). Similarly, the orthogonal regularization can be applied on weights of all these grouped layers.

\section{Perceptual Diversity Metric}
Quantitative metrics for the disentanglement are mostly proposed in VAE-based works and for simulated toy datasets with available ground truth information. \citeauthor{higgins2017beta}~\shortcite{higgins2017beta} suggest training a low-capacity linear classifier on the obtained latent representations of the simulated data from the trained encoder, and report the error rate of the classifier as the disentanglement score of the generative model. \citeauthor{kim2018disentangling}~\shortcite{kim2018disentangling} argue that the introduction of an extra classifier could lead to undesirable uncertainties due to the increased hyper-parameters to tune. Thus, they favor a majority-vote classifier that is achievable directly from the latent representations.

We concur with \citeauthor{kim2018disentangling}~\shortcite{kim2018disentangling} in arguing that, to the best of our knowledge, there is no convincing metric for disentanglement on a dataset for which no ground truth latent factor is provided. Therefore, we propose a method that is capable of relatively evaluating partial properties of a disentangling model when certain conditions are satisfied.

Our intuition is that if a generative model is well-disentangled, then varying each dimension of the controlling vector $c$ should yield different feature changes of the generated data $x$. Suppose the feasible value range for $c$ is $[a, b]$, and for a pair of ($c^o, i, j$) where $c^o$ is a uniformly sampled vector and $i$ and $j$ are two randomly selected indices, we get $c^i$ by setting $c^o[i]=b$ and $c^o[j]=a$, and $c^j$ by setting $c^o[j]=b$ and $c^o[i]=a$. Given the fact that $i$ and $j$ each control different factors, we expect $x^i=G(c^i)$ and $x^j=G(c^j)$ to be different. Therefore, we can use a pre-trained VGG \cite{simonyan2014very} model $V$ to extract the feature map of $x^i$ and $x^j$, and report their $L_1$ distance as the disentanglement score, with a higher $L_1$ distance indicating dimensions $i$ and $j$ are more independent. The final score of this proposed \textit{perceptual diversity metric} will be the average score of many samples of paired ($c^o, i, j$). A formalized algorithmic procedure can be found in the appendix.

\begin{figure}
    \centering
    \includegraphics[width=\linewidth]{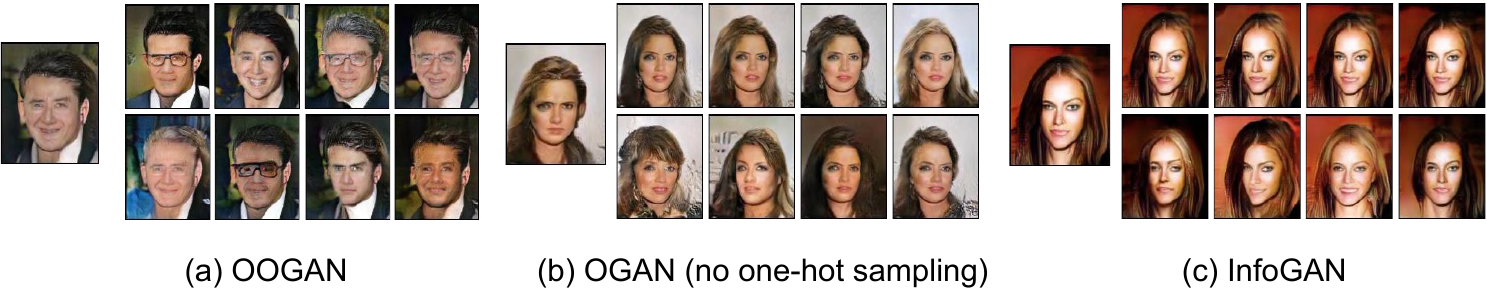}
    \caption{Generated images for CelebA: In each group, the left-most image is generated from a randomly sampled $c$, and the following ones are generated by changing the value of each dimension in $c$ to 1. (a) OOGAN exhibits greater visual differences among each dimension, reflecting its ability to learn diverse latent factors. (b) Without the proposed one-hot sampling, OOGAN still manages to learn some distinguishable features, reflecting the advantage of its structural design. (c) The 4 top right images show that the learned features for an InfoGAN have a large overlap across the latent dimensions in $c$, lacking proper disentanglement.}
    \label{fig:metric}
\end{figure}

We argue that such a metric can adequately reflect the separability and diversity of the learned factors, especially when used for comparing similarly structured models on high-resolution datasets, where higher diversity should already be considered better, and on datasets in which latent factors are known to control a good amount of visual differences. As shown in Figure~\ref{fig:metric}, the proposed metric can efficiently capture the disentangle performance in terms of how diversified each dimension is in $c$. 

\noindent\textbf{Limitations:} The perceptual diversity metric should not be generally used to compare differently structured models, and cannot solely capture the disentanglement ability of a model. First, the L1 distance between feature maps is not an absolute measure. For example, a VAE model $A$ produces blurry images that could lead to a lower value from this metric compared to a GAN model $B$, where images are sharp and high-contrast, but this does not necessarily imply that $A$ disentangles worse than $B$. 

Hence, we only use this metric on the CelebA dataset for comparisons within GAN models, where the aforementioned drawbacks are irrelevant. In such cases, this metric remains valuable in providing an intuitive and direct measure of how well the generative part of the models disentangle. 

\section{Experiments}
We conduct quantitative and qualitative experiments to demonstrate the advantages of our method on several datasets. First, we perform quantitative experiments on the dSprites datasets \cite{dsprites17} following the metric proposed by \citeauthor{kim2018disentangling} \shortcite{kim2018disentangling}. After that, we show the superiority of OOGAN in terms of generating quality images while maintaining competitive disentanglement compared to VAE-based models on CelebA \cite{liu2015faceattributes} and 3D-chair \cite{Aubry14} data. Based on the disentangling benchmark guidance from \citeauthor{eastwood2018framework} \shortcite{eastwood2018framework}, we also present an elaborated learned-factor identification experiment to showcase the effectiveness of OOGAN and validate our compete-and-conflicting issue observations. Finally, we conduct an ablation study on the proposed components in OOGAN with our metric. All the model structures, training details, and more qualitative results are given in the Appendix, and all the code for our experiments is also submitted.

\noindent\textbf{Quantitative results on dSprites:} Several quantitative metrics have been proposed on the dSprites dataset \cite{higgins2017beta,kim2018disentangling,eastwood2018framework,NIPS2018_7527}. While these metrics achieve a thorough evaluation of the disentanglement abilities of the feature-extractor (i.e., the encoder in VAE and $Q$ in GANs), they pay no attention to the generative part of the models. Therefore, we only select \citeauthor{kim2018disentangling}'s metric for its intuitiveness and simplicity to demonstrate our model's competitiveness on the feature extractor's end.

\begin{figure}
    \centering
    \includegraphics[width=\linewidth]{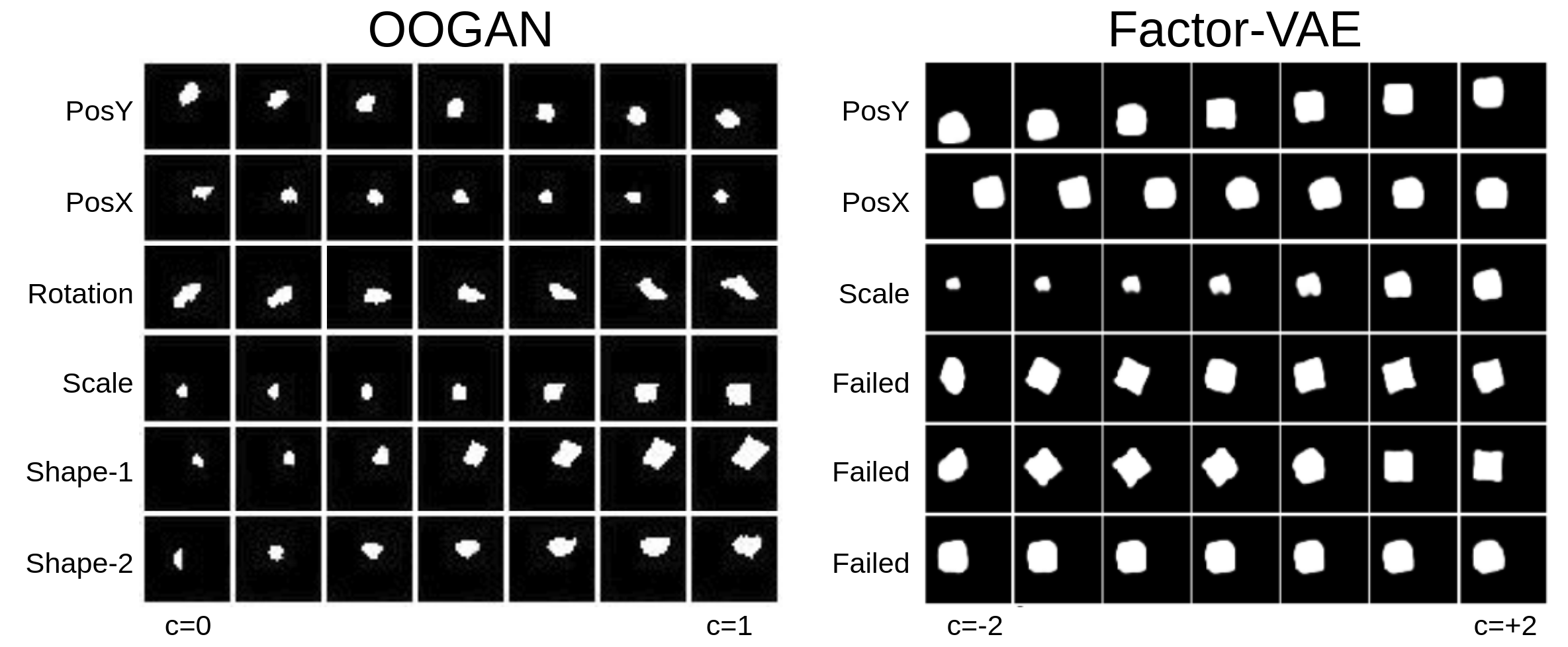}
    \caption{Latent traversals on dSprites}
    \label{fig:dsprite}
\end{figure}

For all the models, we follow the same setup as \citeauthor{kim2018disentangling}~\shortcite{kim2018disentangling} and \citeauthor{jeon2019ibgan}~\shortcite{jeon2019ibgan}. Due to the simplicity of the dataset, we train all the GAN models with the ``instance noise technique" introduced by \citeauthor{sonderby2016amortised}~\shortcite{sonderby2016amortised} to get stable and good quality results.

As can be seen from Table~\ref{table:metrics_kim} and Figure~\ref{fig:dsprite}, our proposed OOGAN genuinely does a better job on both the feature extractor and generator parts. While Factor-VAE is only able to disentangle three out of the five ground truth factors effectively, OOGAN retrieves all the generative factors and manages to put the variables of the discrete factor ``shape" into different dimensions. Additionally, we would like to highlight the robustness of our model, where varying the hyper-parameters of $\lambda$ (1 to 5) and $\gamma$ (0.2 to 2) in our loss function always yields consistent performance. 

\begin{figure}
    \centering
    \includegraphics[width=\linewidth]{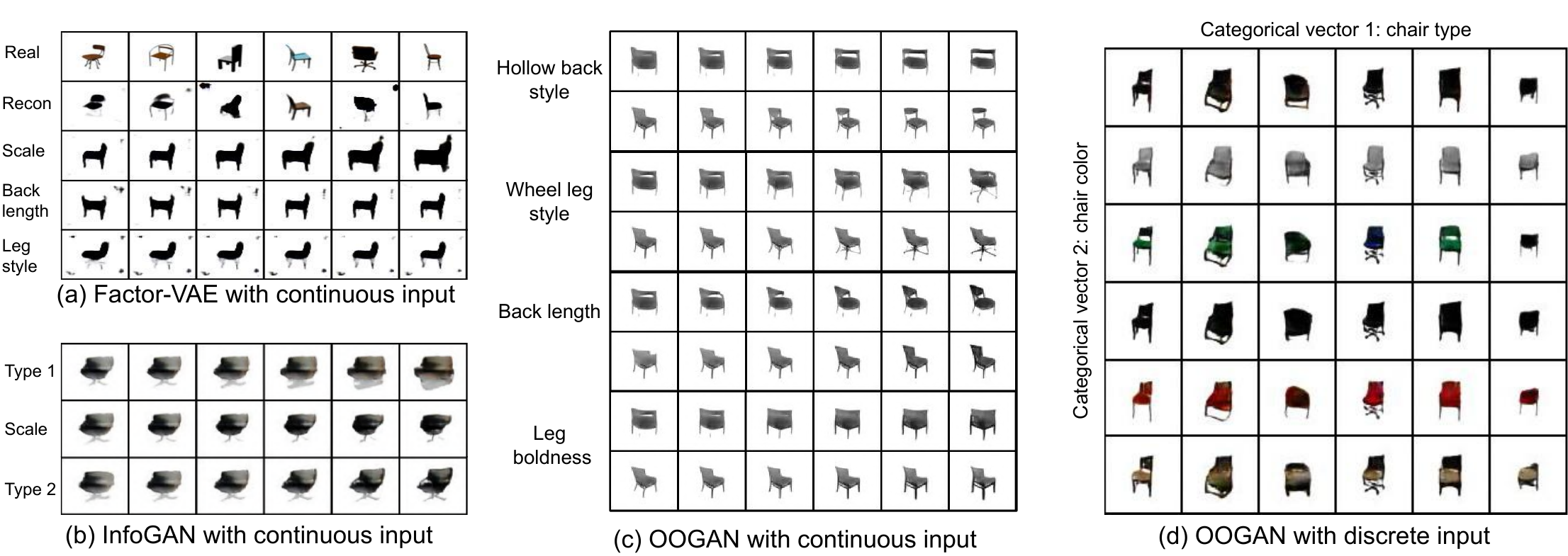}
    \caption{ Latent traversals on 3D Chair}

    \label{fig:chair}
    \vspace{-1.6em}
\end{figure}

\noindent\textbf{Qualitative results on 3D Chairs:}
On the 3D Chairs data, we use $64\times64$ RGB images with batch size 64 for all training runs. To demonstrate the robustness and performance of OOGAN in generating higher-quality images and potentially learning more latent factors, we consider the dimensionality of $c$ for up to 16, where previous works only experiment on a smaller dimensionality such as 6.

In Factor-VAE, when we increase the dimensionality of $c$ from 6 to 16, it struggles to disentangle at a similar quality, and the reconstruction ability is severely sacrificed. In contrast, our model is not affected by an increase of the dimensionality, and apart from learning somewhat more obvious features such as scale and azimuth, our model also discovers several exciting features that have never been reported in previous work. For example, Figure~\ref{fig:chair}-(c) shows a linear transformation of different back styles and leg thickness of the chairs, and Figure~\ref{fig:chair}-(d) shows that our model successfully disentangles discrete features such as ``color" and ``chair type" without any additional tweaks and tricks, for which additional tweaks such as various approximation approaches would have to be incorporated in a VAE approach. 

\begin{table}
    \centering
    \caption{Disentanglement using Kim \emph{et al.}'s metric}

 \begin{tabular}{l|l}
 \hline
  Model       & Score \\
 \hline
  $\beta$-VAE & 0.63 $\pm$0.033 \\
  Factor-VAE  & 0.73 $\pm$0.112 \\
  InfoGAN     & 0.59 $\pm$0.078 \\
  IB-GAN       & 0.80 $\pm$0.062 \\
  OOGAN      & 0.81 $\pm$0.077 \\
\hline
  \end{tabular}
\label{table:metrics_kim}
\end{table}

\begin{table}
    \centering
    \caption{Disentanglement using Perceptual Diversity metric}
 
 \resizebox{0.9\columnwidth}{!}{
 \begin{tabular}{l|l|l}
 \hline
  Model & Score &  Cos-simil.\ in $Q$\\
  \hline
 InfoGAN                      & 2.39 $\pm$0.03            & 0.21 $\pm$ 0.01  \\
  OOGAN w/o One-hot & 2.44 $\pm$0.05            & 0.09 $\pm$ 0.03             \\
 OOGAN w/o Ortho-reg     & 2.65 $\pm$0.05  & 0.21 $\pm$ 0.01 \\
  OOGAN w/o Compt-free G        & 2.69 $\pm$0.03     & 0.09 $\pm$ 0.03         \\
  OOGAN                       & 2.77 $\pm$0.06             & 0.09 $\pm$ 0.03    \\ 
 \hline
  \end{tabular}}
\label{table:metrics_pd}
\end{table}

\noindent\textbf{Disentangling at a higher resolution on CelebA:}
We consider OOGAN as a suite of three modules that can be plugged into any GAN frameworks, and it is orthogonal to other disentanglement approaches based on GANs such as IBGAN. In other words, it can be incorporated into other methods and inherits the breakthroughs made in GANs \cite{zhang2018self,miyato2018spectral,karras2018style}. Therefore, we focus on demonstrating the advantages OOGAN has over VAE-based models in qualitative experiments. The comparison with InfoGAN will be presented as quantitative results in an ablation study.

\begin{figure}
    \centering
    \includegraphics[width=1\linewidth]{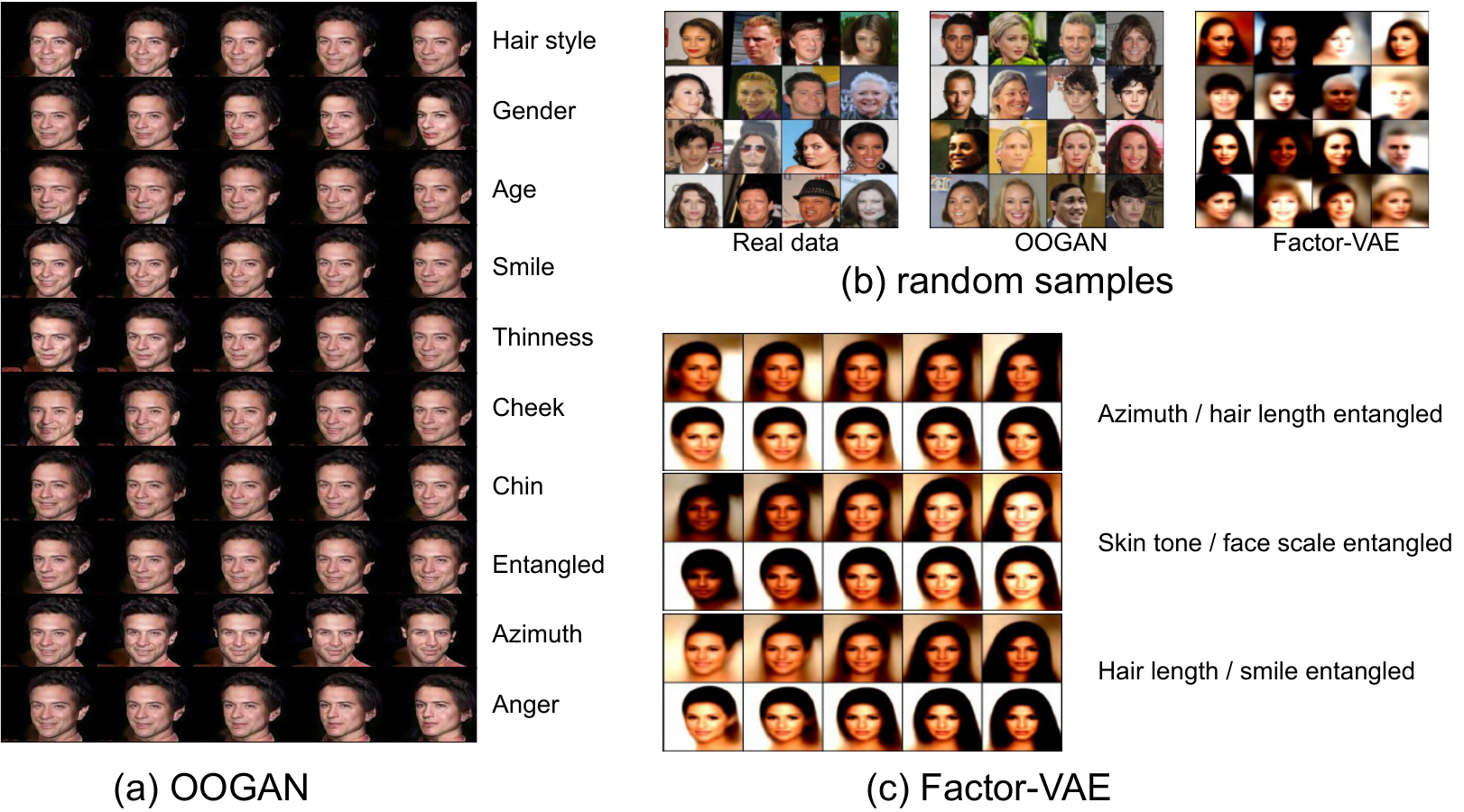}
    \caption{Latent traversals trained on CelebA} 
    \label{fig:CelebA}
\end{figure}

On the CelebA dataset, while previous work operates at a resolution of $64\times64$, we train all the models on the resolution of $256\times256$ to showcase the advantage of OOGAN and expose a shortcoming of the VAE-based models. Figure~\ref{fig:metric}-(a) shows the images trained in a plain DCGAN manner \cite{radford2015unsupervised} and Figure~\ref{fig:CelebA}-(a) shows the images trained in a progressively up-scaling manner \cite{karras2017progressive}, demonstrating a strong ability to disentangle while maintaining a high image quality. On the other hand, VAE-based models deteriorate when reconstructing high-contrast images, 
and are unable to maintain the same disentangle performance as the resolution increases. Thanks to the detail richness of the generated images, OOGAN can discover more interesting facial features such as ``chin" and ``cheek", which no VAE-based models have achieved.

\noindent\textbf{Learned attribute analysis:} To provide a more transparent breakdown on what is learned \cite{eastwood2018framework}, we train 40 binary classifiers on the 40 provided visual attributes from the CelebA dataset, each only predicting one attribute. Then we use these classifiers to monitor the generated images across the training iterations of InfoGAN and OOGAN.  

\begin{figure}
    \centering
    \includegraphics[width=1\linewidth, height=2.3cm]{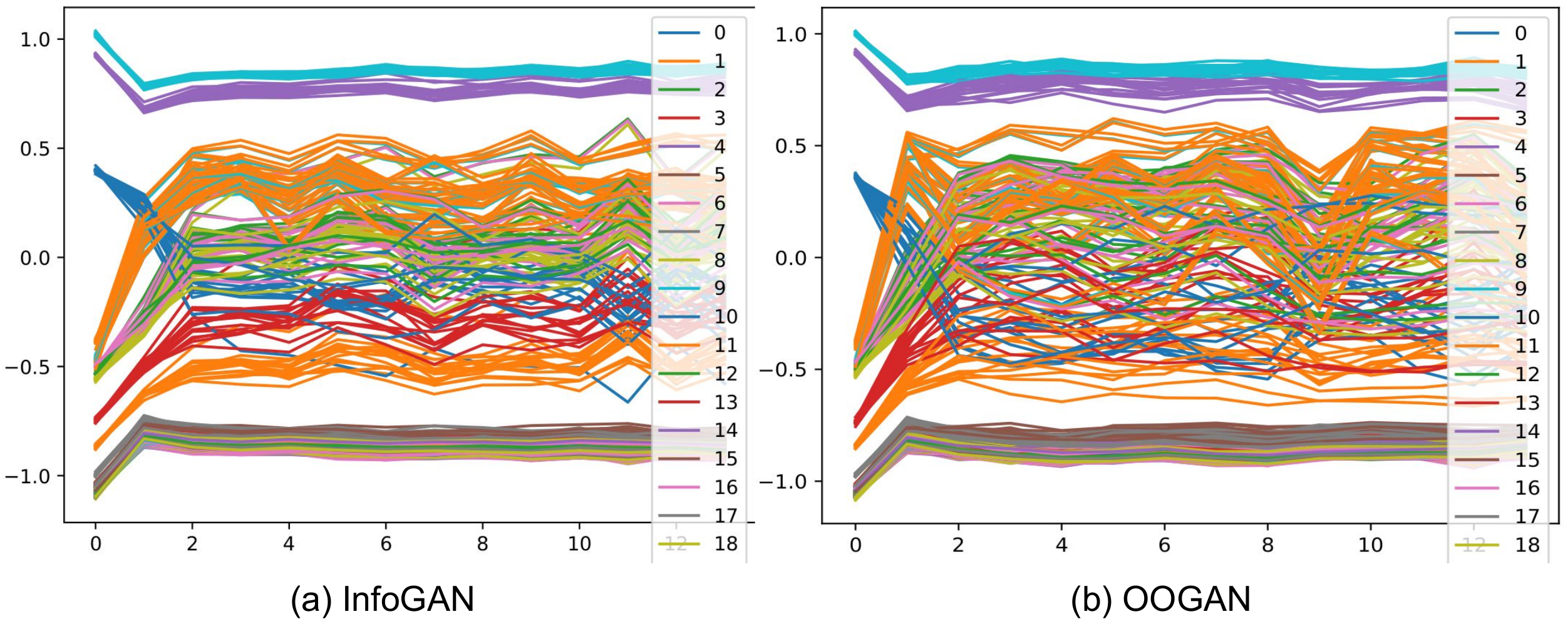}
    \caption{Binary classification of each labeled attributes for each dimension}
    \label{fig:binary_all_dim}
\end{figure}

In Figure~\ref{fig:binary_all_dim}, there are 40 different colors on the lines, each color representing an attribute, with 16 lines per color representing the 16 dimensions in $c$. In terms of sampling $c$, we set one dimension's value to one and sample the remaining dimensions from $\mathsf{uniform}(0, 1)$, repeating the same operation for all dimensions. For InfoGAN, the lines with the same color stick together and have the same tendency to change, which means different dimensions in $c$ are learning similar factors. In contrast, for OOGAN, we observe that lines in the same color (the same attribute) develop differently during the iterations. Moreover, at each iteration, the lines in the same color get different prediction scores, which implies that different dimensions in $c$ are learning different factors.  

\begin{figure}
    \centering
    \includegraphics[width=1\linewidth, height=2.6cm]{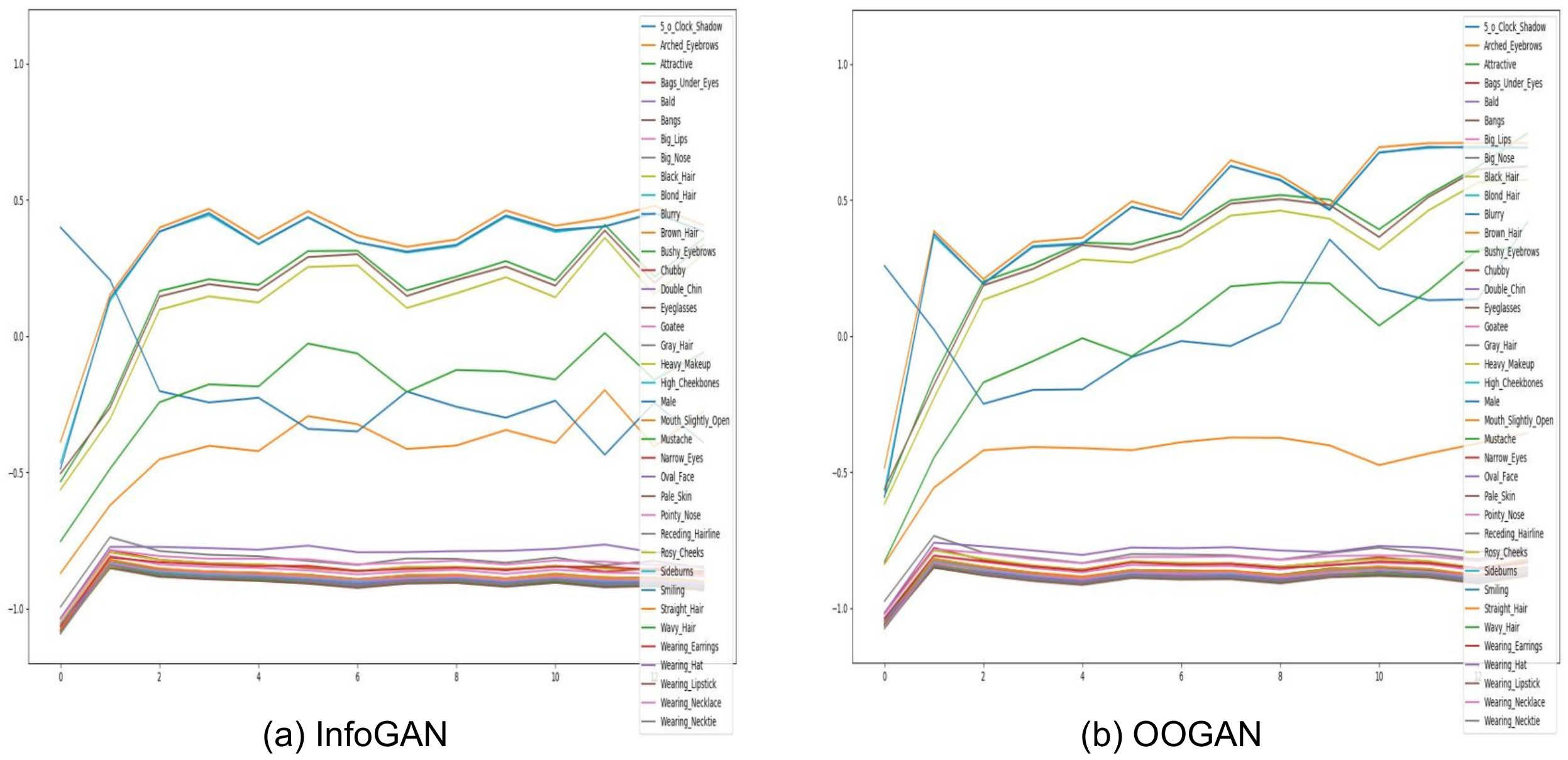}
    \caption{Binary classification of each labeled attributes for all dimensions}
    \label{fig:binary_attri}
\end{figure}

We then average the curves of each attribute into one to show the prediction score for each ground truth attribute in Figure~\ref{fig:binary_attri}, which confirms the \textbf{competing and conflicting issue}. If a prediction score is constantly increasing, this means that $c$ is improving its ability to represent the respective attribute. For InfoGAN, the attribute predictions show rising and falling fluctuations, a sign of an unstable learning process. In contrast, OOGAN steadily increases the prediction score for some visual attributes.

\noindent\textbf{Ablation studies:} Based on the plain InfoGAN setting, we conduct ablation studies on the effectiveness of our proposed three modules quantitatively using our proposed metric, with InfoGAN as the baseline. The experiments are conducted only on learning continuous factors, as InfoGAN already performs well in disentangling categorical latent variables.

There are two types of $Q$ we can choose from when training OOGAN or InfoGAN. A \textit{deterministic $Q$} will try to directly output $c'$, which is considered as a reconstruction of $c$; and a \textit{probabilistic} $Q$ will assume that each dimension of $c_i$ is from a Gaussian and try to output the mean and standard deviation of that distribution given different input images. To optimize a deterministic $Q$, we can directly minimize the L1 loss between predicted $c'$ and the input $c$, and to optimize the probabilistic $Q$, we can minimize the negative log-likelihood given the sampled $c$ and predicted $\mu$ and $\sigma$. In both cases, our proposed one-hot sampling trick can participate in the optimization directly, where for the probabilistic $Q$, we just minimize the cross-entropy between $\mu$ and $c$. 

\begin{figure}
\centering
  \includegraphics[width=\linewidth, height=2.8cm]{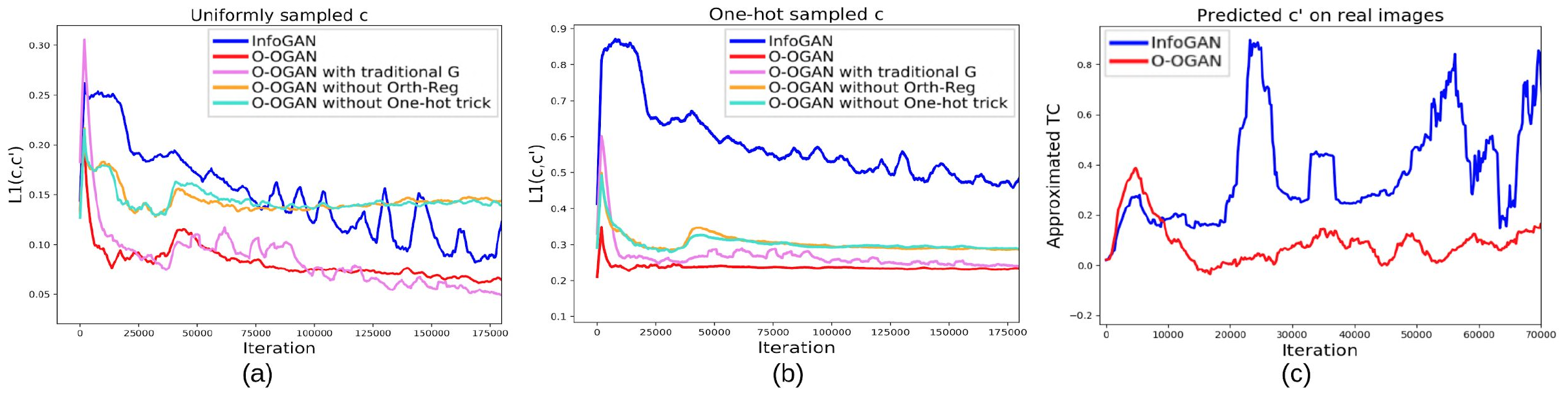}
  \caption{ (a)$, $(b): L1 losses between sampled $c$ and predicted $c'$. (c) TC estimation during training}
  \label{fig:l1loss}
\end{figure}

When the dimensionality of the latent factors $c$ goes beyond 6, InfoGAN fails to disentangle them, which is also confirmed in previous work \cite{kim2018disentangling,jeon2019ibgan}. As shown in Figure~\ref{fig:metric}, most dimensions in InfoGAN produce similar images as many features remain entangled. In the meantime, OOGAN has a better tendency to learn disentangled representations thanks to its structural design, and a direct objective to learn independent features driven by the proposed alternating one-hot sampling.

For the proposed perceptual diversity metric, we fine-tune the VGG model on the CelebA dataset with the provided 40 facial attributes, to make it more sensitive to the visual attributes. 

As shown in Table~\ref{table:metrics_pd}, the one-hot sampling makes the most substantial contribution, while orthogonal-regularized $Q$ and compete-free generator also provide significant improvements. The averaged cosine similarity among the weights is effectively minimized with the proposed orthogonal regularization. In Figures~\ref{fig:l1loss}-(a) and \ref{fig:l1loss}-(b), we train the models with a deterministic $Q$ that directly attempts to reconstruct $c$, and plot the L1 distance between the sampled $c$ and predicted $c'$ (the L1 distance for one-hot $c$ is not used as an objective loss to train InfoGAN). Note how InfoGAN's L1 loss is similarly minimized when $c$ is uniformly sampled, but struggles to decrease when $c$ is one-hot, which means that the output $c'$ of InfoGAN is highly correlated (there are correlated latent factors encoded into multiple dimensions, implying poor disentanglement), while OOGAN’s $c'$ is not. In Figure~\ref{fig:l1loss}-(c), we train the models with probabilistic $Q$ and estimate TC following the method from \citeauthor{NIPS2018_7527}~\shortcite{NIPS2018_7527}. The TC from InfoGAN remains high, while OOGAN can maintain a low TC consistently, which shows the effectiveness of our method. 

\section{Conclusion}
We propose a robust framework that disentangles even high-resolution images with high generation quality. Our one-hot sampling highlights the structural advantage of GANs for easy manipulation on the input distribution that can lead to disentangled representation learning, while the architectural design provides a new perspective on GAN designs. Instead of tweaking the loss functions (designing a new loss, adjusting loss weights, which are highly unstable and inconsistent across datasets), we show that sampling noise from multiple distributions to achieve disentanglement and interpretability is robust and straightforward. It leads to a promising new direction of how to train GANs, opening up substantial avenues for future research, e.g., choosing what distribution to sample from, allocating alternating ratios, etc. The impact of this goes  beyond disentangling, as future research can also be conducted on model interpretability and human-controllable data generation. In the future, we plan to explore more dynamic and fluent sampling methods that can be integrated into the GAN framework for better performance, and we will attempt to validate the benefits of these sampling methods theoretically. 

{\small
\fontsize{9pt}{10pt} \selectfont
\bibliographystyle{aaai}
\bibliography{refs}
}

\begin{center}
	\textbf{\Large OOGAN: Disentangling GAN with\\ One-Hot Sampling and Orthogonal Regularization}
\end{center}

\appendix

\section{OOGAN Training Algorithm}
The complete training process of the proposed OOGAN model is described in Algorithm~\ref{alg:1}.
\FloatBarrier
\begin{algorithm}
\caption{OOGAN training algorithm}
\label{alg:1}
\SetKwInOut{Input}{Input}\SetKwInOut{Output}{Output}
\Input{generator $G$, discriminator $D$, feature extractor $Q$, \\batch size $B$, real data $X$, iteration $n$, control vector dimension $d$.}
\Output{well trained $G$, $D$, $Q$}
\BlankLine
\For{$i$ in $n$ iterations}{
    $x_{real} \gets X$ \;
    $z \gets \mathcal{N}(0, 1)$ \;
    \eIf{$i$ is odd}{
      $indices \gets randint(d)$ \;
      $c \gets onehot(indices)$ \;
    }
    {
      $c \gets uniform(0, 1)$ \;
    }
    $x_{fake} \gets G(z, c)$\;
    \BlankLine
    \BlankLine
    $loss_d \gets relu[1-D(x_{real})] + relu[1+D(x_{fake})]$ \;
    update $D$ via $loss_d$ \;
    \BlankLine
    $loss_g \gets -D(x_{fake})$ \;
    update $G$ via $loss_g$ \;
    \BlankLine
    $loss_{MI} \gets L_1(Q(x_{fake}), c)$ \;
    \If{$i$ is odd}{
        $loss_{MI} \gets loss_{MI} + L_{cross-entropy}(Q(x_{fake}), c)$ \;
    }
    update $G, Q$ via $loss_{MI}$ \;
  }

\end{algorithm}

\section{Perceptual Diversity Metric}
The complete process for the proposed Perceptual-diversity metric is formalized in Algorithm~\ref{alg:2}.

Karras \emph{et al.} \cite{kim2018disentangling} in their StyleGAN work propose the Perceptual Path Length criterion, which is a pairwise image difference between two DNN embeddings of any small cut along with interpolation, and a minor change between the cuts indicates more focused information. However, such a perceptual pairwise distance only reflects the linearity of the interpolated images but cannot determine if a linear change entangles multiple factors or not. Additionally, it has a failure case that when there is no change along with the interpolation, this dimension will get the best score, but actually, nothing is learned.

\begin{algorithm}[t]
\caption{Metric: Perceptual diversity L1 difference}
\label{alg:2}
\SetKwInOut{Input}{Input}\SetKwInOut{Output}{Output}
\Input{generative model $G$, pre-trained VGG model $V$, sample times $n$, latent dimension $d$, latent variable range [$-k$, $k$].}
\Output{score $s$}
\BlankLine
$score = 0$\;
\For{$n$  iterations}{
    Sample $c \in \mathbb{R}^d$ from $Uniform(0,1)$\;
    Sample $i, j$ from $RandomInt$(0, $d$)\;
    $c_1$ = $c$\;
    $c_2$ = $c$\;
    $c_1[i]=-k, c_1[j]=k$\;
    $c_2[i]=k,  c_2[j]=-k$\;
    $feat_1 = V(G(c1))$\;
    $feat_2 = V(G(c2))$\;
    $score = score + L1_{distance}(feat_1, feat_2)$
  }
\Return $score / n$\;
\end{algorithm}

On the other hand, our proposed Perceptual Diversity metric measures how each dimension in $c$ encodes different features that will reflect on the generated images. Moreover, in our experiment on celebA data, we fine-tune the VGG model on the celebA dataset with the provided 40 attributes describing the visual features, to make the VGG model better extract the on-point features, thus making the final score more meaningful.

Note that our proposed metric involves computing the paired L1 difference between two dimensions, while setting one dimension to the maximum value $b$, we will also set the value on the other dimension to the minimum value $a$, to better highlight the visual differences between the dimensions. Since a $c^o$ is uniformly sampled, the original value in $c^o$ on $i$ and $j$ dimension could already be high and close to $b$, so to ensure the two samples $c^i$ and $c^j$ are comparable, we have to force the values in one dimension to $a$.

\section{Experiments}

\subsection{GAN Models}

\textbf{Hardware and training conditions:} We perform all the experiments on one NVIDIA RTX 2080Ti GPU with 11GB vrams. For all the experiments, we train the GAN models with hinge loss and apply spectral norm on all the convolutional layers. We also tried adding a gradient penalty loss, but found that it is not necessary. We rely on Adam optimization with the learning rate set to 1e-4 and betas set to (0.5, 0.99). 

For a vanilla GAN, the running time of OOGAN is nearly identical to InfoGAN, since they share similar training schemas, where for 50,000 iterations, it takes both OOGAN and InfoGAN around 2 hours to train on 32x32 images, 4 hours on 64x64 images, and 10 hours on 128x128 images. The training time is similar to VAE-based models, while a Factor-VAE takes only around half the time to train for the same iterations, it usually takes twice as many iterations to converge to a similar generation quality as GANs. 

For progressively up-scaled GANs, due to the structural similarity, we based our OOGAN on a StyleGAN structure. We modify StyleGAN by adding the proposed Compete-free generator input block, the orthogonal-regularized group convolution $Q$, and training with the alternating one-hot trick. It takes around 20 hours to train on 128x128 images.

\textbf{Model structure details:} On the vanilla GAN's generator, apart from the first input layers, we use convolutional layers with kernel size 3 and stride 1 plus a bi-linear upscale with scale factor 2 to achieve the feature map up-scaling. On the Discriminator side, we also use a convolution layer with kernel size 3 and stride 1, plus an average-pooling with kernel 2 stride 2 to simulate down-scaling. We do not use residual blocks in our models. We generally find such a structure tends to produce smoother images and more stable training compared to using convTransposed layers in the generator. Additionally, we consider the choice of layer to be orthogonal to this work. For a fair comparison, we also conducted an experiment on Factor-VAE with similar convolutional layer up-scaling design, but found VAE to not be sensitive to such model structural changes.

Our model structure overview can be found in Table~\ref{table:64x64-oogan}, where $BN$ is short for batch normalization, $lReLU$ is LeakyReLU, $pool$ means average pooling with stride 2 kernel 2, $upsample$ means a bi-linear up-sampling with factor 2. Our model for 128x128 images is constructed in the same way, with just one more conv-layer block on $G$ and $D$.

\textbf{Approximation of TC:} We also calculate the TC during training to see if our proposed model can produce a lower TC that aligns well with other TC-based methods. To calculate TC, we train the GANs with probabilistic $Q$ and using the \textit{Minibatch Stratified Sampling} method proposed by Chen \emph{et al.} The derivation and details can be found in their paper (TC-VAE)'s appendix section C.2.

Since the code for all our experiments, approximations and metrics is also available (\url{https://github.com/paper-anonymity/OOGAN}), all the training details can be found in the code, if not mentioned here.

\begin{table*}
    \caption{64x64 OOGAN for 3Dchair and dSprites}
    \label{table:64x64-oogan}
    \centering
    \begin{tabular}{lll}
        \toprule
        Discriminator      &  Q   &   Generator \\ \hline
        Input 64x64 RGB image  & -                                & Input c, z       \\
        1x1 conv(sd 1, ch 64)  & shared with D                    & 4x4 convTans(sd 1, ch 512) \\
        3x3 conv(sd 1, ch 128) & shared with D                    & 3x3 conv(sd 1, ch 256)       \\ 
        BN, lReLU(0.1), pool   & shared with D                    & BN, lReLU(0.1), upsample \\
        3x3 conv(sd 1, ch 256) & shared with D                    & 3x3 conv(sd 1, ch 256)   \\
        BN, lReLU(0.1), pool   & shared with D                    & BN, lReLU(0.1), upsample \\
        3x3 conv(sd 1, ch 256) & 3x3 orth-group conv(sd 1, ch c)  & 3x3 conv(sd 1, ch 128) \\
        BN, lReLU(0.1), pool   & lReLU(0.1), pool                 & BN, lReLU(0.1), upsample \\
        3x3 conv(sd 1, ch 512) & 4x4 orth-group conv(sd 2, ch c)  & 3x3 conv(sd 1, ch 64)\\
        BN, lReLU(0.1), pool   & lReLU(0.1), pool                 & BN, lReLU(0.1), upsample \\
        4x4 conv(sd 1, ch 1)   & linear layer with output c       & 3x3 conv(sd 1, ch 3), sigmoid \\
        \bottomrule
    \end{tabular}
\end{table*}

\section{More Qualitative Results}

\begin{figure*}
\centering
  \includegraphics[width=1\linewidth]{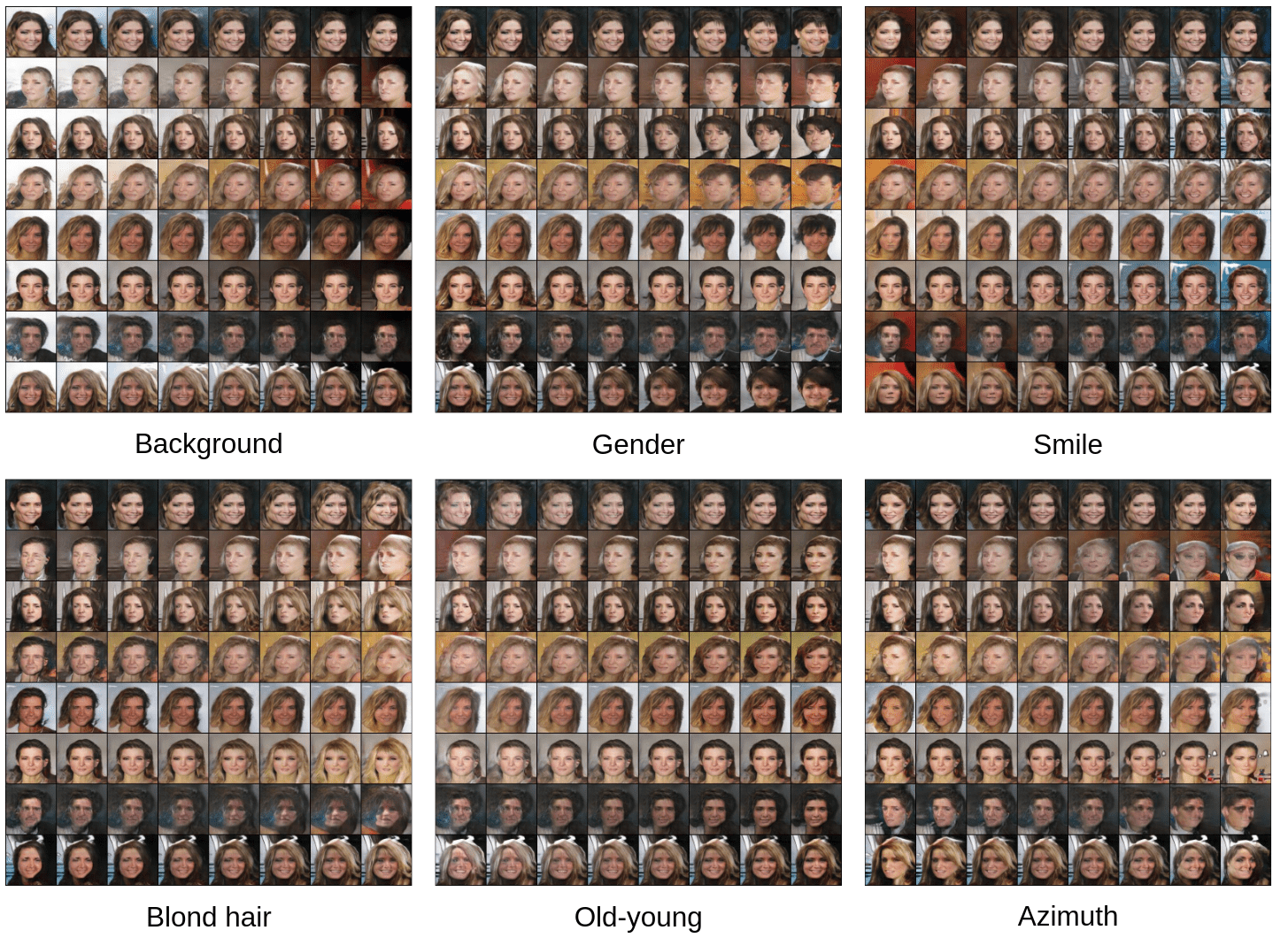}
  \captionof{figure}{ OOGAN based on Vanilla GAN. Latent traversals along different dimensions.}
  \label{fig:celebA_more_vanilla}
\end{figure*}

\begin{figure*}
\centering
  \includegraphics[width=\linewidth]{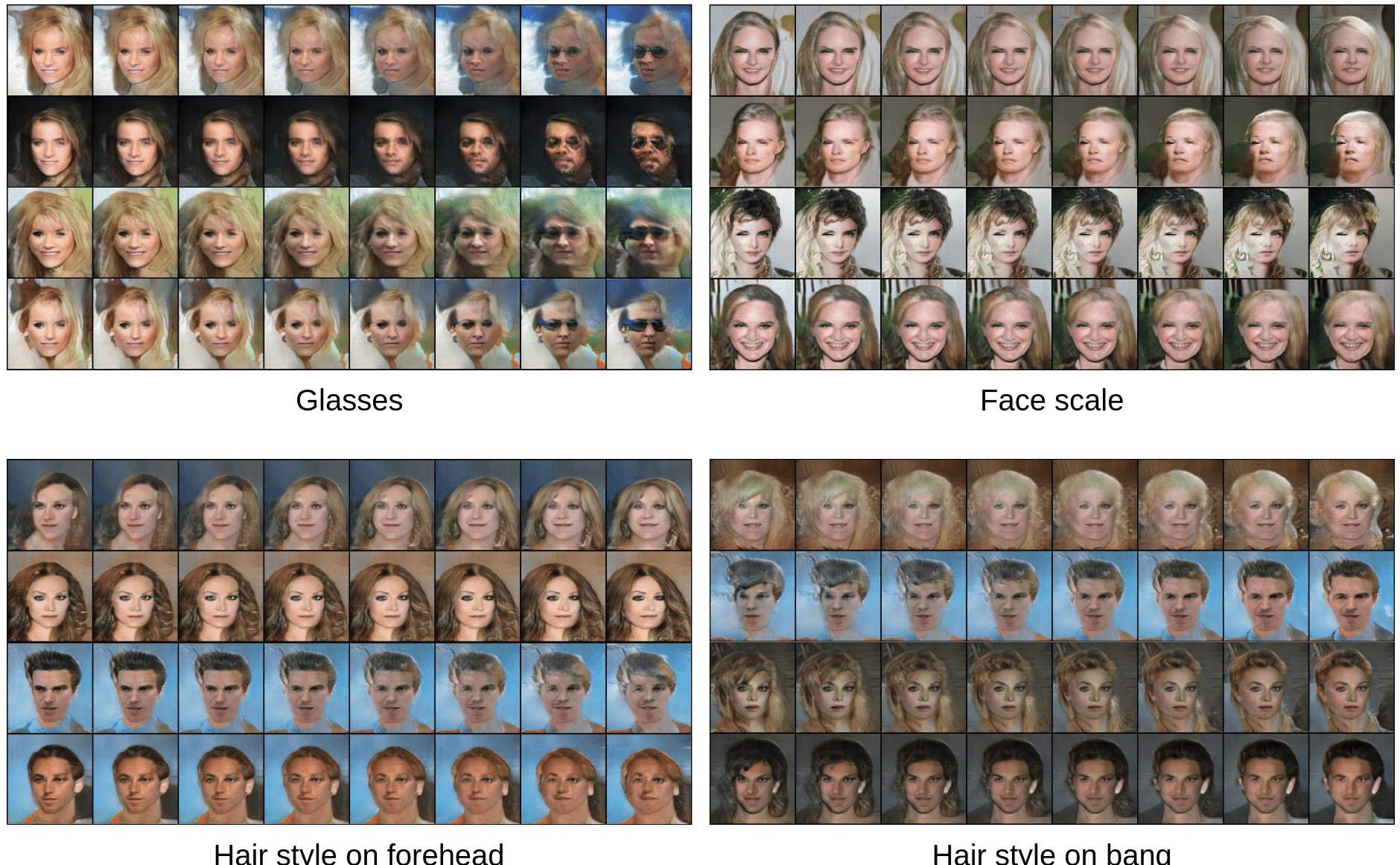}
  \captionof{figure}{ OOGAN based on Vanilla GAN. Latent traversals along different dimensions.}
  \label{fig:celebA_more_vanilla_2}
\end{figure*}

\begin{figure*}
\centering
  \includegraphics[width=\linewidth]{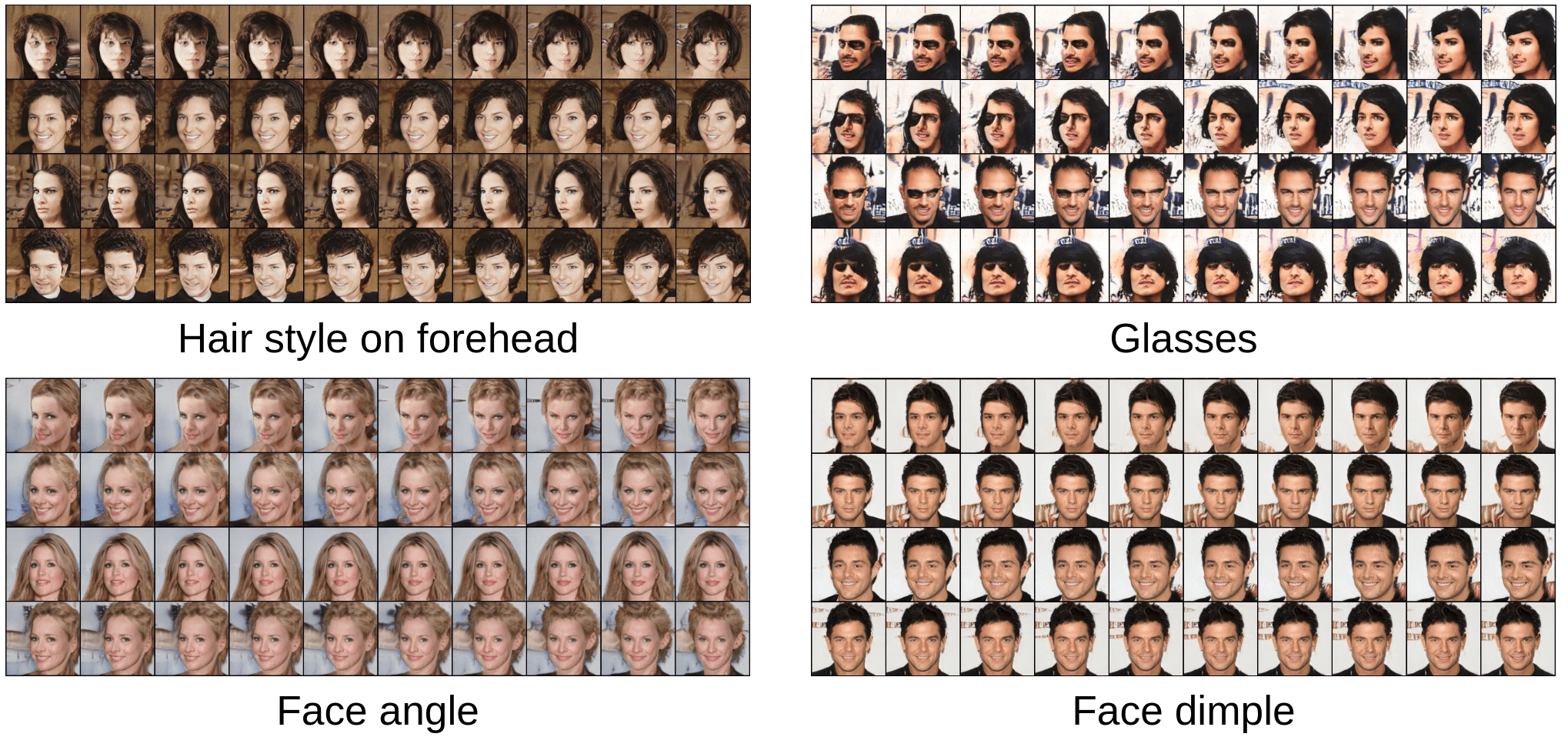}
  \captionof{figure}{ OOGAN based on StyleGAN. Latent traversals along different dimensions.}
  \label{fig:celebA_more_stylegan}
\end{figure*}

\begin{figure*}
\centering
  \includegraphics[width=\linewidth]{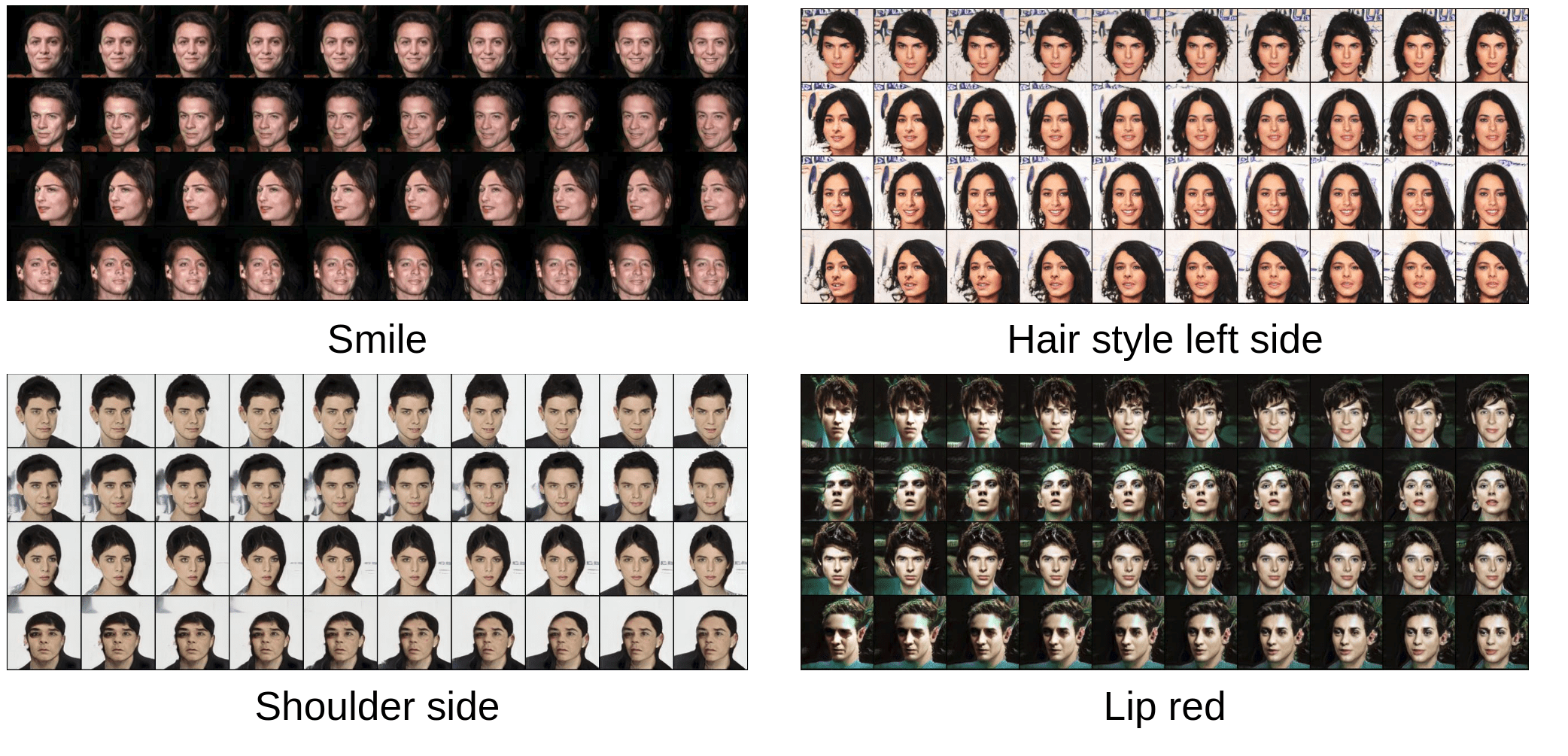}
  \captionof{figure}{ OOGAN based on StyleGAN. Latent traversals along different dimensions.}
  \label{fig:celebA_more_stylegan_2}
\end{figure*}

\begin{figure*}
\centering
  \includegraphics[width=\linewidth]{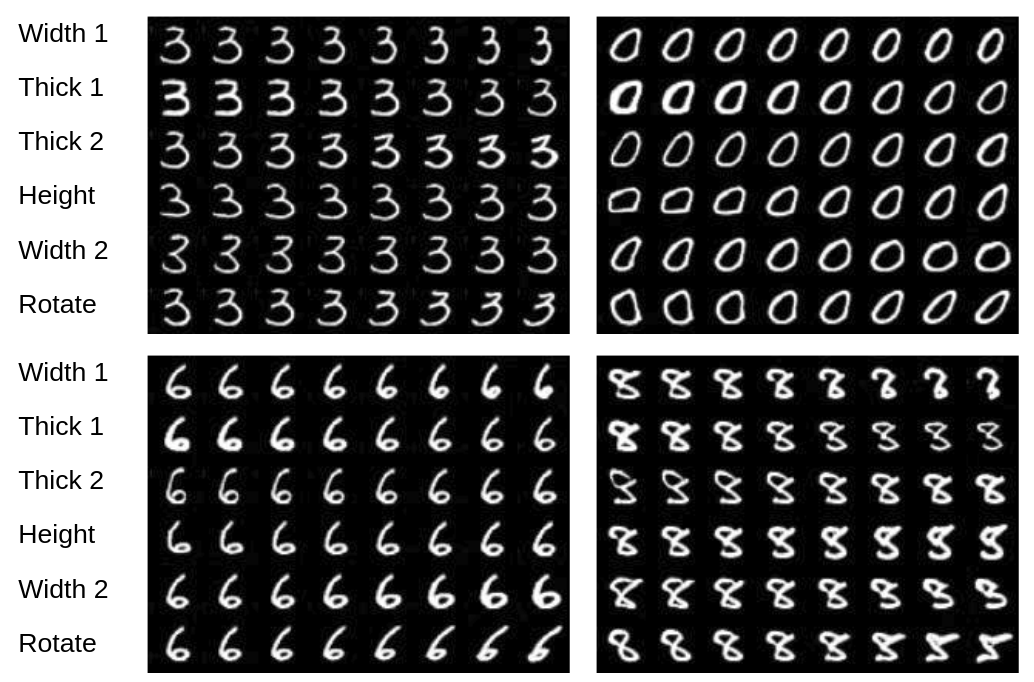}
  \captionof{figure}{ OOGAN on MNIST dataset, the 4 sections are 4 selected indices from the discrete vector (categorical one-hot vector), inside each section are latent traversals on the same 6 continuous dimensions. }
  \label{fig:mnist}
\end{figure*}

\end{document}